\documentclass{article} % For LaTeX2e
\usepackage{iclr2026_conference,times}

\usepackage{amsmath,amsfonts,bm}

\def\eqref#1{equation~\ref{#1}}
\def\1{\bm{1}}

\DeclareMathAlphabet{\mathsfit}{\encodingdefault}{\sfdefault}{m}{sl}
\SetMathAlphabet{\mathsfit}{bold}{\encodingdefault}{\sfdefault}{bx}{n}

\usepackage{hyperref}
\usepackage{url}
\usepackage{graphicx}
\usepackage{tabularx}
\usepackage{algorithm}
\usepackage{algpseudocode}

\usepackage{listings}
\usepackage{xcolor}

\newcolumntype{C}[1]{>{\centering\arraybackslash}m{#1}}

\usepackage{enumitem}

\definecolor{codegray}{rgb}{0.95,0.95,0.95} % 背景灰
\definecolor{codeborder}{rgb}{0.8,0.8,0.8}  % 边框灰
\definecolor{deepblue}{rgb}{0,0,0.6}
\definecolor{deepgreen}{rgb}{0,0.5,0}
\definecolor{deepred}{rgb}{0.6,0,0}

\allowdisplaybreaks[4]

\usepackage{booktabs} % 用于制作专业表格
\usepackage{multirow} % 用于合并单元格
\usepackage[table]{xcolor} % 用于表格着色

\usepackage[caption=false,font=footnotesize,labelfont=rm,textfont=rm]{subfig}

\title{Learning from the Best, Differently: A Diver-sity-Driven Rethinking on Data Selection}

\author{
\textbf{Hongyi He\textsuperscript{1}\textsuperscript{*}}\ \ \ 
\textbf{Xiao Liu\textsuperscript{2}\textsuperscript{\dag}}\ \ \ 
\textbf{Zhenghao Lin\textsuperscript{2}}\ \ \ 
\textbf{Mingni Tang\textsuperscript{3}\textsuperscript{*}}\ \ \ 
\textbf{Yi Cheng\textsuperscript{3}\textsuperscript{*}}\ \ \ \\
\textbf{Jintao Wang\textsuperscript{1}\textsuperscript{\dag}}\ \ \ 
\textbf{Wenjie Li\textsuperscript{3}}\ \ \ 
\textbf{Peng Cheng\textsuperscript{2}}\ \ 
\textbf{Yeyun Gong\textsuperscript{2}\textsuperscript{\dag}}\ \\
\textsuperscript{1} Tsinghua University, \textsuperscript{2} Microsoft Research, \textsuperscript{3} The Hong Kong Polytechnic University\\
\texttt{hehy22@mails.tsinghua.edu.cn}, \\
\texttt{\{xiaoliu2,zhenghaolin,pengc,yegong\}@microsoft.com}, \\
\texttt{wangjitao@tsinghua.edu.cn}, \\
\texttt{\{minnie17.tang,alyssa.cheng\}@connect.polyu.hk}, \\
\texttt{cswjli@comp.polyu.edu.hk} \\
}

\iclrfinalcopy % Uncomment for camera-ready version, but NOT for submission.
\begin{document}

\renewcommand{\thefootnote}{\fnsymbol{footnote}}
% \footnotetext[1]{Equal contribution.}
\footnotetext[1]{Work done during their internships at Microsoft Research.}
\footnotetext[2]{Corresponding authors.}
\renewcommand{\thefootnote}{\arabic{footnote}}  % optional

\maketitle

\begin{abstract}
High-quality pre-training data is a decisive factor for large language models, where quality captures factual reliability and semantic value, and diversity ensures broad coverage and distributional heterogeneity. Existing approaches typically rely on single or multiple-dimensional score-based selection. However, empirical studies have shown that directly selecting top-scored data often degrades downstream performance, and sampling from a broader range is required to recover results. The above non-monotonicity between the dataset scores and the downstream benchmark results reveals a fundamental bias: score-based methods collapse correlated dimensions, causing top-scored data to appear high-quality while systematically overlooking diversity. We argue that ensuring diversity requires decomposing correlated evaluation metrics into orthogonal feature dimensions, from which the top-scored data can be directly selected. To this end, we proposed the \textbf{O}rthogonal \textbf{Di}versity-Aware \textbf{S}election (\textbf{ODiS}) algorithm, a method to preserve both quality and diversity during high-quality data selection. First, ODiS evaluates data from multiple dimensions, covering language quality, knowledge quality, and comprehension difficulty. The resulting multi-dimensional scores are then decorrelated via Principal Component Analysis (PCA), yielding orthogonal evaluation dimensions. For each dimension, a Roberta-based scorer is trained to regress the data onto PCA-projected scores, enabling scalable inference on large corpora. Finally, ODiS constructs the training dataset by selecting top-scored data within each orthogonal dimension, thereby ensuring both quality and diversity.  Empirical results show that ODiS-selected data exhibit less than 2\% inter-dimension overlap, confirming the orthogonality between dimensions. More importantly, models trained with ODiS-selected data significantly outperform other baselines on multiple downstream benchmarks, highlighting the necessity of orthogonal, diversity-aware data selection for LLMs.
\end{abstract}

\section{Introduction}
Pretraining is the primary stage for models to acquire fundamental abilities, such as language understanding, text generation and information extraction (\cite{brown2020language, chowdhery2023palm, roberts2020much}). These capabilities are largely determined by the quality and diversity of the training data. Quality captures authenticity, reliability, and semantic integrity, ensuring that models learn accurate and well-structured knowledge. Diversity, on the other hand, emphasizes coverage and comprehensiveness, enabling models to generalize across domains and tasks. With the increase of both model and corpus sizes, designing efficient data selection methods that jointly account for these two aspects has become a critical challenge for advancing model performance.

Existing works have proposed diverse methods for selecting data based on quality and diversity. Quality-based methods typically utilize rule-based heuristics (\cite{laurenccon2022bigscience, weber2024redpajama, penedo2023refinedweb, raffel2020exploring, lee2021deduplicating}), such as document length constraint and content deduplication, or score-based techniques (\cite{wenzek2019ccnet, touvron2023llama, wettig2024qurating, penedo2024fineweb, su2024nemotron}), where classifiers or perplexity models assign a single quality score to filter noisy or irrelevant data. Diversity-based methods (\cite{he2024softdedup, zhang2024harnessing, tirumala2023d4, yang2025diversity}) instead focus on broadening coverage by mixing data across domains or clustering in the embedding space, thereby reducing redundancy and expanding the distribution. Recent works also attempt to combine quality and diversity to select data (\cite{zhuang2025meta, liu2025quadmix, bai2025efficient}), typically formulating both diversity and quality into a multi-dimensional score or mixing selected data from different domains. However, the intrinsic correlation between dimensions makes weight tuning challenging, and naive aggregation often results in overlapping signals. In summary, current works encounter three key challenges: \textbf{(1)} The data selected with top scores are not always optimal, and sampling is necessary to achieve satisfactory performance, but the cause remains unexplored. \textbf{(2)} Scored-based methods combine multiple aspects into a one-dimensional signal, making them unable to capture both quality and diversity. \textbf{(3)} Delicate hyper-parameter tuning is required to balance the influence from different dimensions, undermining generality and practical deployment of the methods.

To address these challenges, we first revisit the problem of data selection through the lens of bias and correlation, identifying the neglect of diversity as the fundamental cause. Guided by the insight, we propose \textbf{O}rthogonal \textbf{Di}versity-Aware \textbf{S}election (ODiS) algorithm, a method to effectively construct a dataset with both quality and diversity. Specifically, drawing inspiration from (\cite{zhuang2025meta, wettig2024qurating}), we label a reference dataset from 11 dimensions, covering four main categories: language quality, knowledge quality,  comprehension difficulty, and information quality. After analyzing the correlation across the dimensions, we reveal strong entanglement between them, which will introduce redundancy and bias the score-based selection. To mitigate this, we apply Principal Component Analysis (PCA) to scores and derive orthogonal evaluation dimensions. For acceleration and scalability, Roberta-based scorers are trained to predict scores along each PC dimension on the target dataset. Finally, ODiS constructs the training set by selecting the top samples from orthogonal PC dimensions, ensuring both quality and diversity of the dataset.

Empirical validations demonstrate that the model trained on data selected by the proposed ODiS methods achieves the best in various downstream benchmarks, compared with existing baselines DSIR, PPL, and Nemotron-CC. We analyze the source of performance gains by comparing top-scored data with samples drawn from broader ranges of each dimension, and find that top-scored subsets consistently underperform, whereas combining data across dimensions substantially improves performance. Data analysis further confirms the strong dimension correlations before PCA and demonstrates that orthogonal principal components capture distinct aspects of the data, thereby validating the data quality and diversity in the selected dataset. Finally, the ablation study suggests that increasing the number of dimensions will marginally improve performance, indicating an efficiency-performance trade-off.

The main contributions of this work are as follows: \textbf{(1)} We provide the first analysis about performance degradation of top-scored data, identifying neglected diversity as the underlying cause. \textbf{(2)} We propose the ODiS algorithm, a score-based data selection algorithm that explicitly ensures both quality and diversity through dimension decomposition.  \textbf{(3)} The analysis results reveal that ODiS benefits from dimension decomposition and enhanced data diversity, which sheds light on future data selection methods for reducing inter-dimensional correlations to improve data diversity. 

% The empirical experiments validate the performance of the proposed method in enhancing data diversity and model training performance.

\section{Method}
\subsection{Score-based Bias in Data Selection\label{sec:bias-analysis}}
We begin by analyzing the causes of selection bias and the role of data sampling through examining model performance with varying data sizes. Following \cite{wettig2024qurating}, we establish multiple metrics targeting different semantic features. Using these metrics, we utilize a score-based method to select a subset of data from the Nemotron-CC dataset (\cite{su2024nemotron}). Specifically, we filter the top-scored data at different scales, ranging from 100B to 900B tokens. After that, we train a 1.5B-parameter model from scratch on these subsets, with 100B tokens training data budget. Finally, model performance is evaluated across multiple benchmarks. The details of the evaluation metrics, training details, and benchmark selection are provided in Sections \ref{sec:multi-dimensions}, \ref{sec:experiment-setup}, and \ref{sec:experiment-setup}, respectively.
    \begin{figure}[ht]
    \centering
    \vspace{-0.3cm}
        \subfloat[Arc-Easy]{
            \centering
            \label{fig:average-performance1}
            \includegraphics[width=0.18\linewidth]{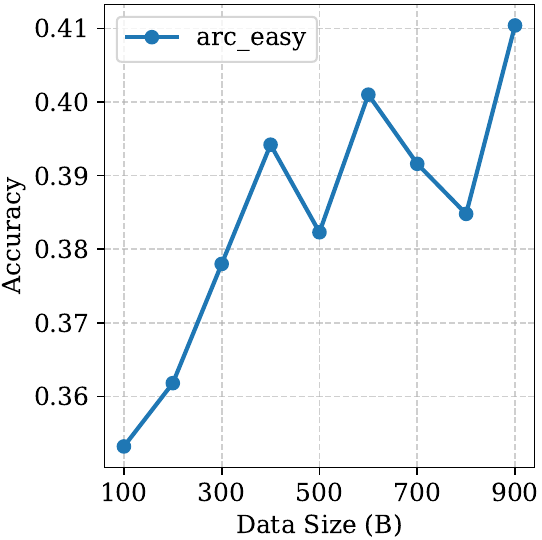}
        }
        \hspace{0.03cm}
        \subfloat[Arc-Challenge]{
            \centering
            \label{fig:average-performance2}
            \includegraphics[width=0.18\linewidth]{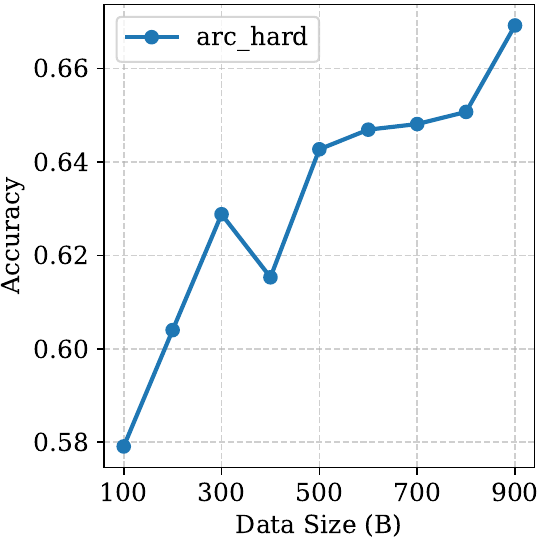}
        }
        \hspace{0.03cm}
        \subfloat[Hellaswag]{
            \centering
            \label{fig:average-performance3}
            \includegraphics[width=0.18\linewidth]{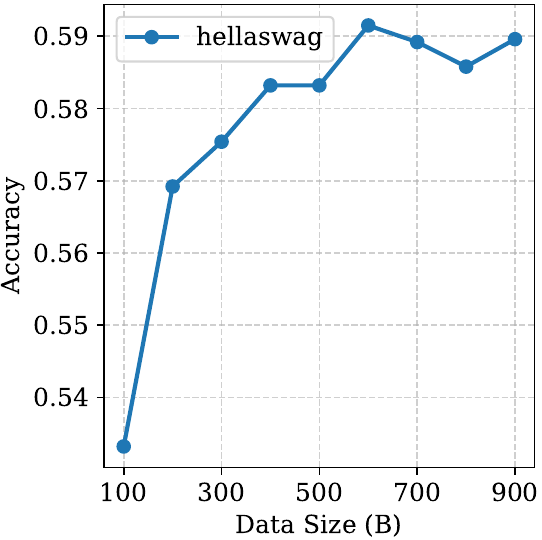}
        }
        \hspace{0.03cm}
        \subfloat[PIQA]{
            \centering
            \label{fig:average-performance4}
            \includegraphics[width=0.18\linewidth]{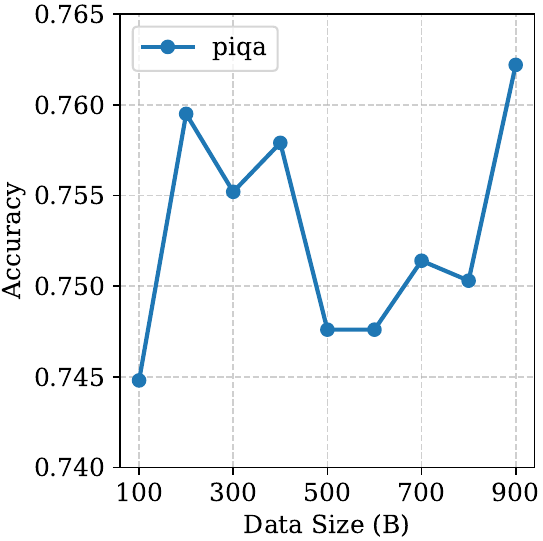}
        }
        \hspace{0.03cm}
        \subfloat[SCIQ]{
            \centering
            \label{fig:average-performance5}
            \includegraphics[width=0.18\linewidth]{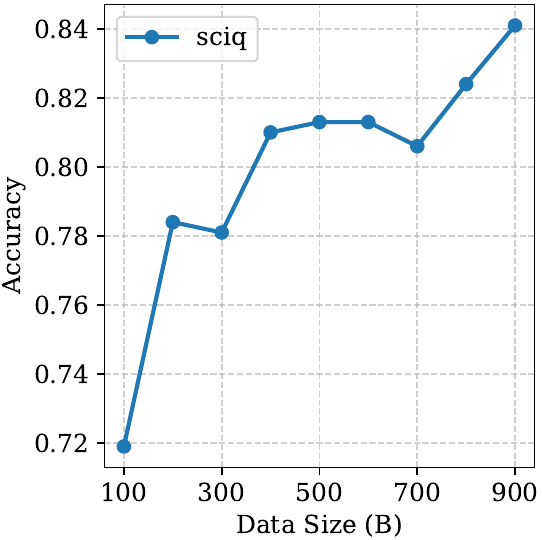}
        }
        \caption{Performances across downstream tasks.}
        \label{fig:average-performance}
        \vspace{-0.5cm}
    \end{figure}
From Figure \ref{fig:average-performance}, we can observe that the data with the highest score performs the worst, while sampling data from a broader range leads to improvements. This reveals a non-monotonic relationship between the data score and model performance, which complicates data selection: the sampling range and other potential hyperparameters should be optimized accordingly. Since the top-scored data has the highest quality, we further investigate its diversity through an embedding-based visualization to determine the reason behind the non-monotonicity. 

\begin{figure}[ht]
    \centering
    \vspace{-0.3cm}
        \subfloat[UMAP projection of text embeddings. The embeddings are reduced to two dimensions using UMAP.]{
            \centering
            \label{fig:diversity-umap}
            \includegraphics[width=0.37\linewidth]{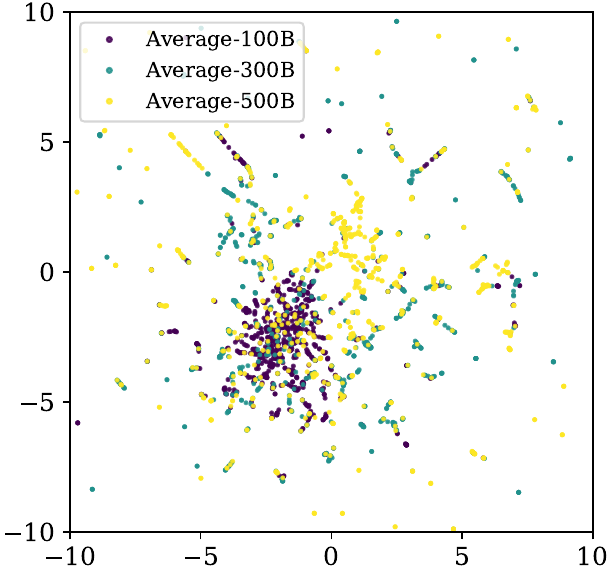}
        }
        \hspace{0.2cm}
        \subfloat[Pairwise distance distribution of text embeddings. The distances are computed with cosine distances.]{
            \centering
            \label{fig:diversity-distance}
            \includegraphics[width=0.47\linewidth]{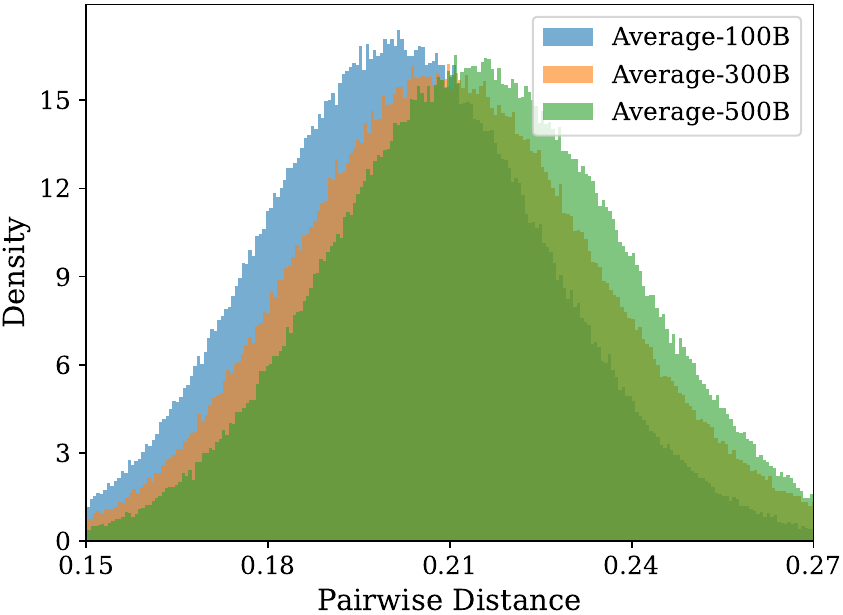}
        }
        \caption{Visualization of data distribution from different data pool sizes. The embeddings are generated using 1000 randomly sampled texts with the m3e-base model, followed by L2 normalization and removal of the top 3 principal components to suppress dominant common directions. }
        \label{fig:diversity-analysis}
        \vspace{-0.5cm}
    \end{figure}

We sample the data from different sizes of data pools, and visualize the UMAP projection of text embeddings together with the distribution of pairwise distances. As shown in Figure \ref{fig:diversity-umap}, a larger data subset spreads more broadly in the compressed dimension. Moreover, Figure \ref{fig:diversity-distance} demonstrates that the pairwise distance increases with the data size. These results suggest that the increase in data size widens the distinction between data and amplifies data diversity, whereas top-scored data is relatively homogeneous, which explains the performance degradation of the top-scored data.

\subsection{Problem Formulation}
Motivated by the previous analysis in Section \ref{sec:bias-analysis}, we focus on enhancing data diversity while ensuring data quality during data selection. Our objective is to select the most valuable subset of a large corpus to facilitate the model pre-training. Instead of directly processing the target dataset $\mathcal{D}_t$, we first take a smaller reference dataset $\mathcal{D}_r$ to generate the scoring strategy. Specifically, each data $x_i$ in the reference dataset $\mathcal{D}_r=\{x_i\}_{i=1}^N$ will be labeled from $m$ dimensions, whose score vector can be expressed as $\boldsymbol{\alpha}^{(i)} = (\alpha_1^{(i)}, \alpha_2^{(i)}, \cdots, \alpha_{m}^{(i)})$. The goal is to design a mapping function $F(\cdot)$ that transforms the scores $\boldsymbol{\alpha}^{(i)}$ into a ranking, through which we can directly select the top-scored data. The function should be designed such that training on the top-scored data maximizes the performance of downstream tasks. The problem can be written as follows:
\begin{equation}
    \mathcal{D}_s = \arg \max_{\mathcal{D}_s \subset \mathcal{D}_t} G(\mathcal{D}_s),
\end{equation}
where $G(\cdot)$ denotes evaluation performance on benchmarks tasks. 

\subsection{Multi-Dimension Data Evaluation \label{sec:multi-dimensions}}
Instead of directly optimizing the data selection for downstream tasks, which may be biased toward task-specific signals, we propose a method that focuses on enhancing data quality and diversity for general purposes. To comprehensively evaluate each data document, we set up 11 dimensions, i.e., $m=11$, for four general aspects: Language quality, Knowledge quality, Comprehension difficulty, and Information Quality. Without loss of generality, we still use $m$ to denote the number of dimensions in the algorithm. We briefly describe the dimensions as follows, and the details are in Appendix \ref{appd:11-dimension}:

\begin{itemize}[leftmargin=1em]
\item \textbf{Language quality.}
We prefer data that is (i) coherent in structure, (ii) concise without redundancy, and (iii) correct in spelling/grammar and word choice (\cite{penedo2023refinedweb}).

\item \textbf{Knowledge quality.}
We value content with (i) sufficient coverage and (ii) depth, (iii) useful reasoning signals, (iv) clear educational, and (v) practical value (\cite{gunasekar2023textbooks,guo2025deepseek}).

\item \textbf{Comprehension difficulty.}
We assess the difficulty level, i.e., conceptual complexity and domain professionalism, as higher difficulty can improve generalization (\cite{agrawal2023corpus}).

\item \textbf{Information quality.}
We require (i) factual accuracy and (ii) sufficient completeness so models can learn reliable, fully specified facts (\cite{chang2024large}).
\end{itemize}

% dimension, metric, PC dimension

Each document is assigned a score from 0 to 5 on every dimension using the OpenAI GPT API. Detailed definitions of the metrics and prompt are provided in Appendix~\ref{appd:score-metric}. The resulting scores constitute a matrix $\mathbf{X}$:
\begin{equation}\mathbf{X}=\begin{bmatrix}\alpha_1^{(1)}&\cdots&\alpha_{m}^{(1)}\\\vdots&\ddots&\vdots\\\alpha_1^{(N)}&\cdots&\alpha_{m}^{(N)}\end{bmatrix}\in \mathbb{R}^{N\times m}.
\end{equation}

\subsection{Orthogonal Diversity-Aware Selection}
\textbf{Dimension decomposition via PCA.} Previous studies have shown that different dimensions often exhibit correlations(\cite{zhuang2025meta}), i.e., the data with higher knowledge depth may have less knowledge richness, while the data with high educational value usually achieves high information quality. Such redundancy in the raw labeled score reduces effective data diversity and hinders data selection. Therefore, instead of directly using the raw labeled scores, we will transform the scores to eliminate the potential correlation between different dimensions, which is done through principal component analysis (PCA).

To eliminate the scale difference, we first calculate the mean of each dimension $\boldsymbol{\mu}$, and obtain the data matrix $\mathbf{X}_c$ centered with the mean:
\begin{equation}
    \boldsymbol{\mu}=\frac{1}{N}\sum_{i=1}^N \boldsymbol{\alpha}^{(i)}, \mathbf{X}_c=\mathbf{X}-\boldsymbol{\mu}.
    \label{eq:mean-calculation}
\end{equation}
After that, we compute the covariance matrix  $\mathbf{\Sigma}$ and adopt eigen decomposition to $\mathbf{\Sigma}$:
\begin{equation}
    \boldsymbol{\Sigma} = \boldsymbol{V} \boldsymbol{\Lambda} \boldsymbol{V}^T, \mathbf{\Sigma}=\frac{1}{N-1}\mathbf{X_c}^T\mathbf{X_c} \in \mathbb{R}^{m\times m},
\end{equation}
where $\boldsymbol{V} = [\mathbf{v}_1, \cdots, \mathbf{v}_{m}]$ are orthogonal eigenvectors and $\boldsymbol{\Lambda}=\text{diag}(\lambda_1 \geq \cdots \lambda_{m} \geq 0)$ are eigenvalues. Each eigenvector represents an orthogonal combination of the original metrics, with the eigenvalue quantifying the variance contribution, i.e., the proportion of the feature representation that the eigenvector accounts for. The higher eigenvalue indicates that the data spreads more along this direction, and the eigenvector contains a more representative feature.

\textbf{Score transformation.} To reduce cost and improve efficiency, we take the first $K$ principal components (PC) to satisfy the explained-variance ratio, rather than all the PCs: $\frac{\sum_{k=1}^{K}\lambda_k}{\sum_{k=1}^{m}\lambda_k} \geq \tau, $  
where $\tau$ is the threshold. Then, we can obtain the project matrix: $\mathbf{W}_K=[\mathbf{v}_1, \cdots, \mathbf{v}_K]\in \mathbb{R}^{m\times K}$. Through projection, the compressed score vector for $x_i$ is calculated as: $\boldsymbol{\beta}^{(i)} = \mathbf{W}_K^T(\boldsymbol{\alpha}^{(i)})^T \in \mathbb{R}^K$. Note that the principal component is a linear combination of the original dimensions, and it is difficult to interpret its meaning. Since the dimensions are orthogonal, we have decomposed each metric, and the score in each principal component represents the quality in this new dimension. 

\textbf{Roberta-enabled model-based scorer.} To enhance labeling accuracy and capture the semantic feature of the principal component, we train a Roberta-based scorer $r_k(\cdot)$ to map the original text to the PCA-derived score for each PC dimension. The scorer $r_k(\cdot)$ will regress from the data $x_i$ from the reference dataset and the transformed score $\beta^{(i)}_k$. The Roberta-based scorer enables efficient inference on unseen data, preserves semantic richness, and provides a unified and noise-robust scoring framework for large-scale data selection. We then label the target dataset $\mathcal{D}_t$ with scorer $r_k(\cdot)$ to obtain the scores $\theta^{(i)}_k, k\in {1, \cdots, K}, i\in \{1, \cdots, |\mathcal{D}_t|\}$

\textbf{Dataset construction based on scores.} We allocate a data budget $s_k$ to each PC dimension considering its contribution and the total data budget: $s=\sum_{k=1}^K s_k$. For each dimension, a score threshold $t_k$ is set based on $s_k$, and the corresponding subset is denoted as  $\mathcal{D}^k_s$. Since labeling and scorer training inevitably introduce noise, the top-scored data across each orthogonal dimension may still overlap, and directly merging the subsets will result in duplication. To address this, we apply a joint score threshold $\boldsymbol{t}=(t_1, \cdots, t_K)$ and select a data within the target dataset $x_i\in\mathcal{D}_t$ if $\theta_k^{(i)}>t_k, \forall k\in {1, \cdots, K}$. The final selected data is obtained as the union: $\mathcal{D}_s=\sum_{k=1}^K \mathcal{D}_s^k$. This construction ensures that each dimension contributes its highest-quality data, while the orthogonality of PC dimensions guarantees enhanced diversity in the resulting dataset.

\begin{algorithm}[t]
\caption{Orthogonal Diversity-Aware Selection}
\label{alg:example}
\begin{algorithmic}[1]  % [1] 表示显示行号
\Require Reference dataset $\mathcal{D}_r$, target dataset $\mathcal{D}_t$, dimensions for evaluation, data budget $s$
\Ensure Selected dataset $\mathcal{D}_s$
\State For each $x_i \in \mathcal{D}_r$, obtain the $m$ dimensional score vector  $\boldsymbol{\alpha}^{(i)} = (\alpha_1^{(i)}, \alpha_2^{(i)}, \cdots, \alpha_{m}^{(i)})$;
\State Compute the mean vector $\boldsymbol{\mu}$ and construct the centered matrix $\boldsymbol{X}_c$ as \eqref{eq:mean-calculation};
\State Compute the covariance matrix $\boldsymbol{\Sigma}$, and perform eigendecomposition to obtain eigenvalues $\lambda_i$ and eigenvectors $\mathbf{v}_i$;
\State Determine the number of principal components (PCs) $K$ with the threshold $\tau$;
\State Construct the project matrix $\boldsymbol{W}_K$, and project scores into the orthogonal space $\boldsymbol{\beta}^{(i)} = \mathbf{W}_K^T(\boldsymbol{\alpha}^{(i)})^T \in \mathbb{R}^K$;
\State Allocate budget $s_k$ to each PC dimension such that $\sum_{k=1}^K s_k = s$;
\For{$k = 1, \dots, K$}
    \State Train a RoBERTa-based scorer $r_k(\cdot)\in [0,5]$ by regressing text $x_i \in \mathcal{D}_r$ to the PCA-transformed scores $\beta_k^{(i)}$;
    \State Apply $r_k(\cdot)$ to obtain predicted scores $\{\theta_k^{(i)}\}$ for all $x_i \in \mathcal{D}_t$;
    \State Given the budget $s_k$, determine threshold $t_k$ and select $\mathcal{D}^k_s = \{x_i \mid \theta_k^{(i)} > t_k, x_i\in\mathcal{D}_t\}$;
    
\EndFor
\State Construct the final selected dataset $\mathcal{D}_s=\cup_{k=1}^{K} \mathcal{D}_s^k$.
\end{algorithmic}
\end{algorithm}

\section{Experiments}
\subsection{Experiment Setup \label{sec:experiment-setup}}
\textbf{Dataset.} We use Nemotron-CC dataset (\cite{su2024nemotron}) as the data pool for selection. It is a large-scale Chinese dataset for pretraining large language models. The dataset comprises both real-world and synthetic data, covering major domains, such as web knowledge and question answering. For tokenization, we use the LLaMA-3-8B tokenizer, which has a vocabulary size of 128,256, and set the maximum sequence length to 4096.

\textbf{Evaluation.} We evaluate the model performance with lm-eval-harness framework (\cite{eval-harness}). We first monitor task-level performance fluctuations across training steps, with detailed results presented in Appendix \ref{sec:benchmarks}. Based on the above result, we follow the "fine task" metric (\cite{kydlicek2024finetasksmultilingualtasks}) to select downstream tasks with performance that varies significantly as training progresses. We select five tasks covering main categories: \textbf{General Knowledge} (including Arc-Easy/Challenge (\cite{clark2018think})), \textbf{Commonsense Reasoning} (including Hellaswag (\cite{zellers2019hellaswag}), SIQA (\cite{sap2019-social})), and \textbf{Physical Reasoning} (PIQA (\cite{bisk2020piqa})).

\textbf{Training.} Each experiment is conducted under a data budget of 100B tokens. Unless specified in the caption, the \textit{top} selection represents directly selecting a 100B token dataset for training, while \textit{sample} selection represents selecting a 700B token dataset and sampling training data from it. We employ a decoder-only model with 1.5 billion parameters and train it from scratch with the selected data. Training uses a global batch size of 512 and the AdamW optimizer (\cite{loshchilov2018decoupled}), with a peak learning rate of 3e-4, cosine decay scheduling, and linear warmup. Unless specified, the proposed ODiS method selects data from the first 4 PC dimensions and allocates the data budget evenly to each dimension.

\subsection{Main Results}
The main results are summarized in Table \ref{tab:main_results}. We compare the proposed ODiS method with existing baselines, including DSIR (\cite{xie2023data}), PPL (\cite{ankner2024perplexed}), and Nemotron-CC (\cite{su2024nemotron}). The results show that the model trained on data selected by ODiS achieves the highest performance compared with the baselines. Specifically, ODiS achieves a generally 3-point marginal improvement compared with the random sampling in average accuracy. Notably, it surpasses all the methods across all task categories, highlighting its versatility in addressing a wide range of downstream tasks. The performance gains are typically obvious in Arc-E and Arc-C, indicating their enhancement in general knowledge and content diversity. In contrast, baseline methods such as PPL and DSIR emphasize only data quality, which limits diversity and hampers overall performance. These results demonstrate the importance of data diversity during data selection, and decomposition is a direction for efficient diversity improvement.
\begin{table}[h]
\centering
\begin{tabular}{l|C{1.3cm} C{1.3cm} C{1.3cm} C{1.3cm} C{1.3cm} C{1.3cm}}
\toprule
\textbf{Method}  &\textbf{Arc-C} &\textbf{Arc-E} &\textbf{Hellaswag} &\textbf{SIQA} &\textbf{PIQA} &\textbf{Average}
\\ \midrule
Random Selection & 0.3503 & 0.6273 & 0.5825 & 0.855 & 0.7448 & 0.6320 \\ 
Nemotron-HQ & 0.3725 & 0.6463 & 0.5774 & 0.839 & 0.7356 & 0.6341 \\ 
PPL-\textit{Top} & 0.3788 & 0.6284 & 0.5469 & 0.834 & 0.7465 & 0.6269 \\
PPL-\textit{Sample} & 0.3609 & 0.6431 & 0.5837 & 0.858 & 0.7481 & 0.6388 \\ 
DSIR & 0.2782 & 0.4848 & 0.5457 & 0.785 & 0.7095 & 0.5606 \\ 
PC Average-\textit{Top} & 0.3532 & 0.6481 & 0.5332 & 0.719  & 0.7448 & 0.5859 \\
PC Average-\textit{Sample} & 0.3916 & 0.5791 & 0.5892 & 0.806  & 0.7514 & 0.6373 \\ \midrule
\textbf{ODiS} & \textbf{0.4155} & \textbf{0.6694} & \textbf{0.5842} & \textbf{0.872}  & \textbf{0.7740} & \textbf{0.6597} \\
\bottomrule
\end{tabular}
\caption{Performance across data selection methods. The random selection method samples data from the whole Nemotron-CC dataset, while the Nemotron-HQ method samples from the Nemotron-CC HQ subset. The PC Average baseline utilizes the averaged scores from PC1 to PC4.}
\label{tab:main_results}
\vspace{-0.5cm}
\end{table}

\subsection{Analysis Result}
% We conduct an analysis study from the perspective of data bias, dimension correlation, data orthogonality, and scaling with more dimensions.
\subsubsection{Inspection of Data Bias} 
\begin{figure}[h]
    \centering
    \vspace{-0.3cm}
    \subfloat[Performance with data selected from PCs]{
            \centering
            \label{fig:performance-top-sample}
            \includegraphics[width=0.40\linewidth]{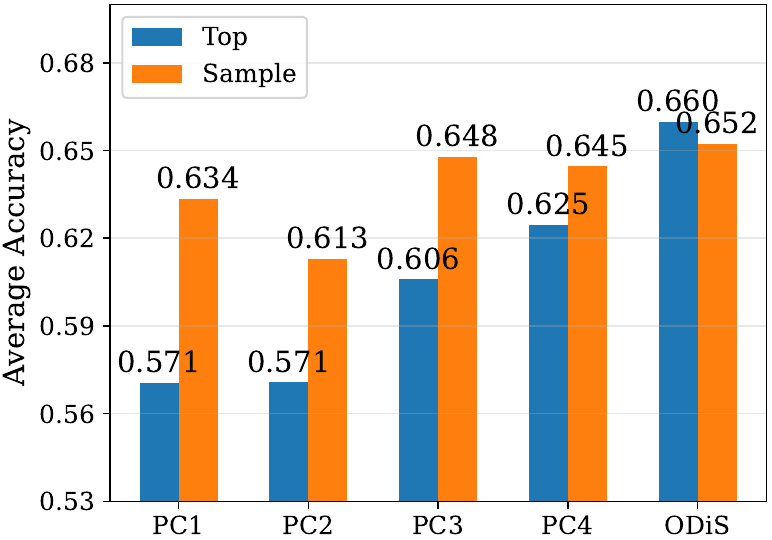}
        }
        \hspace{0.5cm}
        \subfloat[Performance across selected data pools]{
            \centering
            \label{fig:performance-ours-average}
            \includegraphics[width=0.40\linewidth]{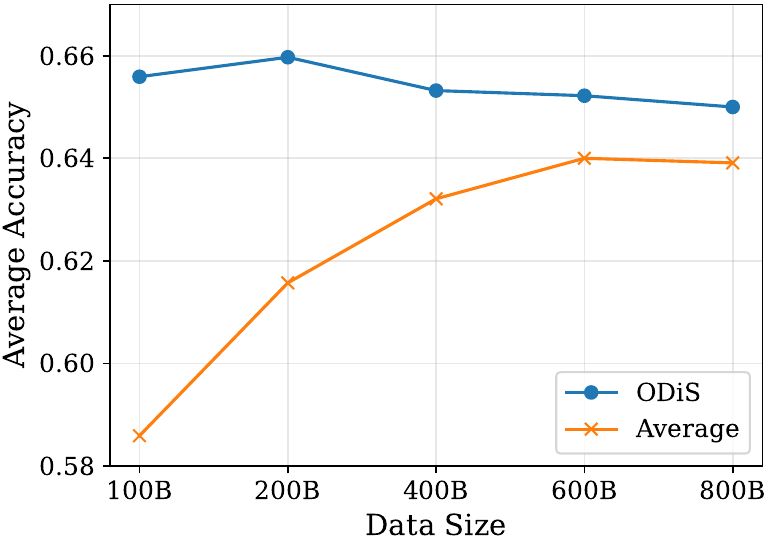}
        }
    \caption{Model performance under different selection methods and ranges.}
    \vspace{-0.3cm}
    \label{fig:performance-bias}
\end{figure}

\textbf{ODiS mitigates data bias.} Figure~\ref{fig:performance-top-sample} shows that models trained with top-scored data within a single dimension consistently underperform, while sampling from a broader score range yields significant performance gains. This observation highlights the inherent bias in relying exclusively on single-dimension scoring, where overemphasis on one metric neglects complementary aspects of data quality and diversity. Furthermore, Figure~\ref{fig:performance-ours-average} illustrates that averaging scores from multiple dimensions only partially alleviates the data bias issue, as correlation across dimensions discussed in Section~\ref{sec:dimension-correlation} continues to bias the data selection. In contrast, ODiS decomposes the dimensions into orthogonal components and selects high-quality data from each dimension, effectively mitigating data bias and enhancing diversity. Notably, ODiS achieves superior performance with the smallest subset of data, avoiding both excessive sampling ranges and unnecessary data waste.

\begin{figure}[h]
    \centering
        \vspace{-0.3cm}
        \subfloat[Pairwise distance distribution]{
            \centering
            \label{fig:pairwise-distance-after}
            \includegraphics[width=0.37\linewidth]{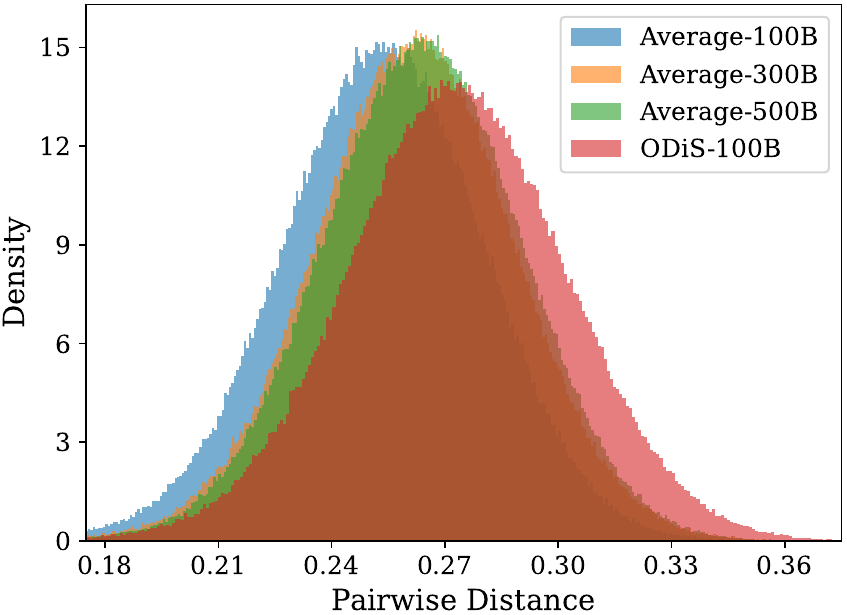}
        }
        \hspace{0.5cm}
        \subfloat[UMAP projection of text embeddings]{
            \centering
            \label{fig:umap-after}
            \includegraphics[width=0.36\linewidth]{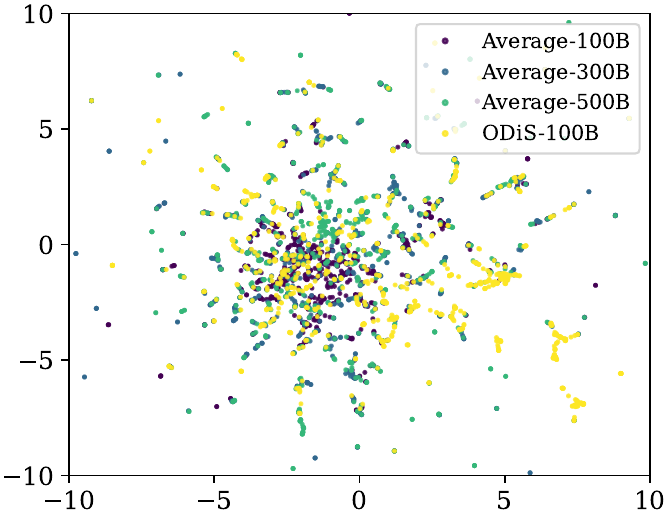}
        }
        \caption{Data diversity visualization from different selection methods.}
        \label{fig:dimension-diversity-after}
        \vspace{-0.5cm}
    \end{figure}

\textbf{ODiS effectively enhances data diversity.} To validate the diversity gain from ODiS, we compare the ODiS-selected data against score-based baselines. Figure \ref{fig:dimension-diversity-after} shows that the data selected by ODiS has a significantly larger average pairwise distance, even compared with a larger selected data pool, indicating enhanced data diversity. Similarly, UMAP visualization reveals that the ODiS-selected data spans a wider region in the compressed space, while the data selected with baseline methods tend to cluster around a narrower region. These results suggest that ODiS reduces redundancy and captures a wider range of semantic features during data selection.

\subsubsection{Correlation between dimensions \label{sec:dimension-correlation}}
% To better understand the performance gain of our method, we conduct data analysis from dimension correlation, feature distribution, and data diversity. First, w

\begin{figure}[h]
    \centering
    \vspace{-0.3cm}
        \subfloat[Correlation matrix between different dimensions]{
            \centering
            \label{fig:correlation_matrix}
            \includegraphics[width=0.57\linewidth]{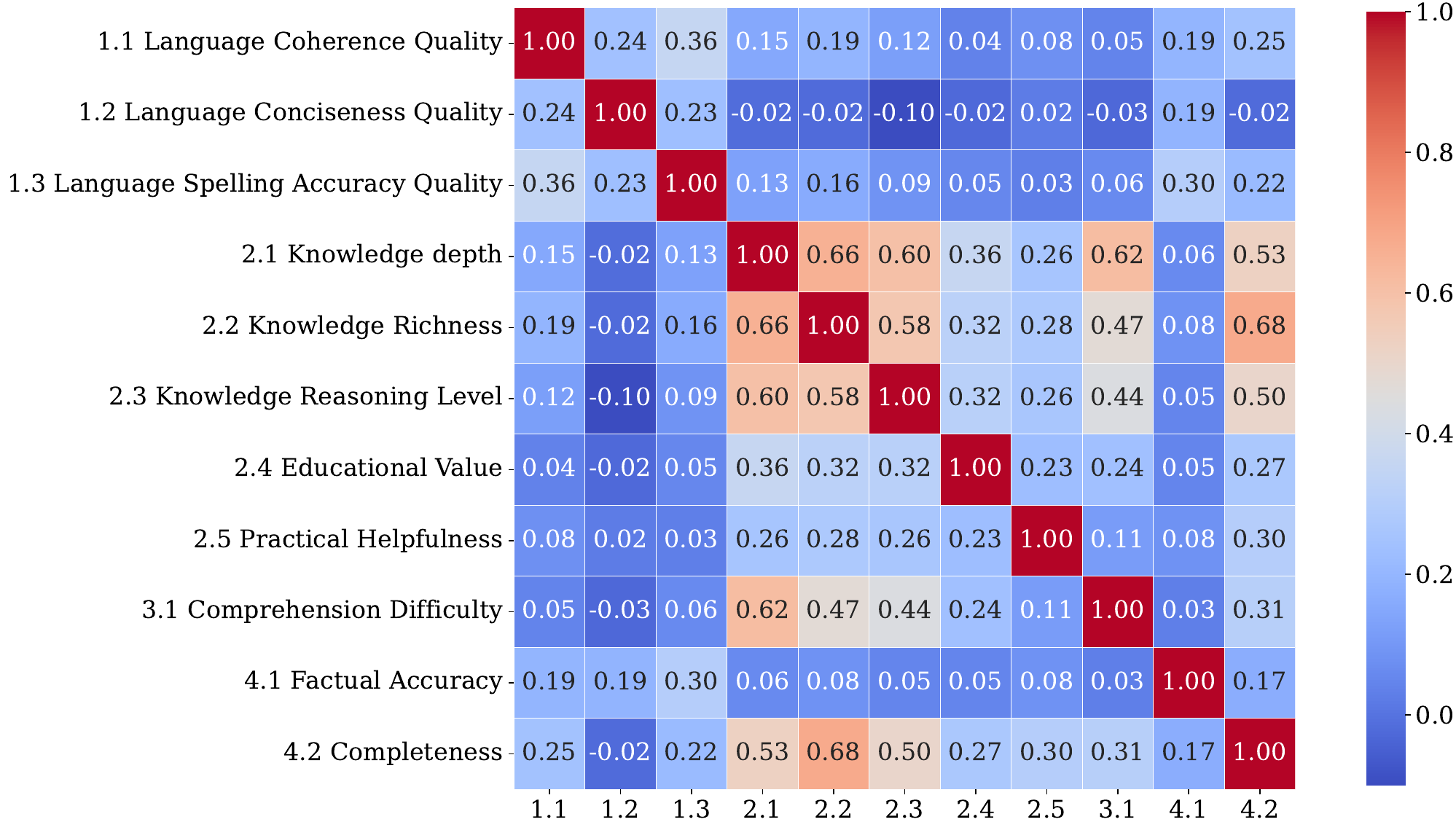}
        }
        \hspace{0.1cm}
        \subfloat[Correlation matrix between original and PC dimensions]{
            \centering
            \label{fig:correlation_matrix_pc}
            \includegraphics[width=0.39\linewidth]{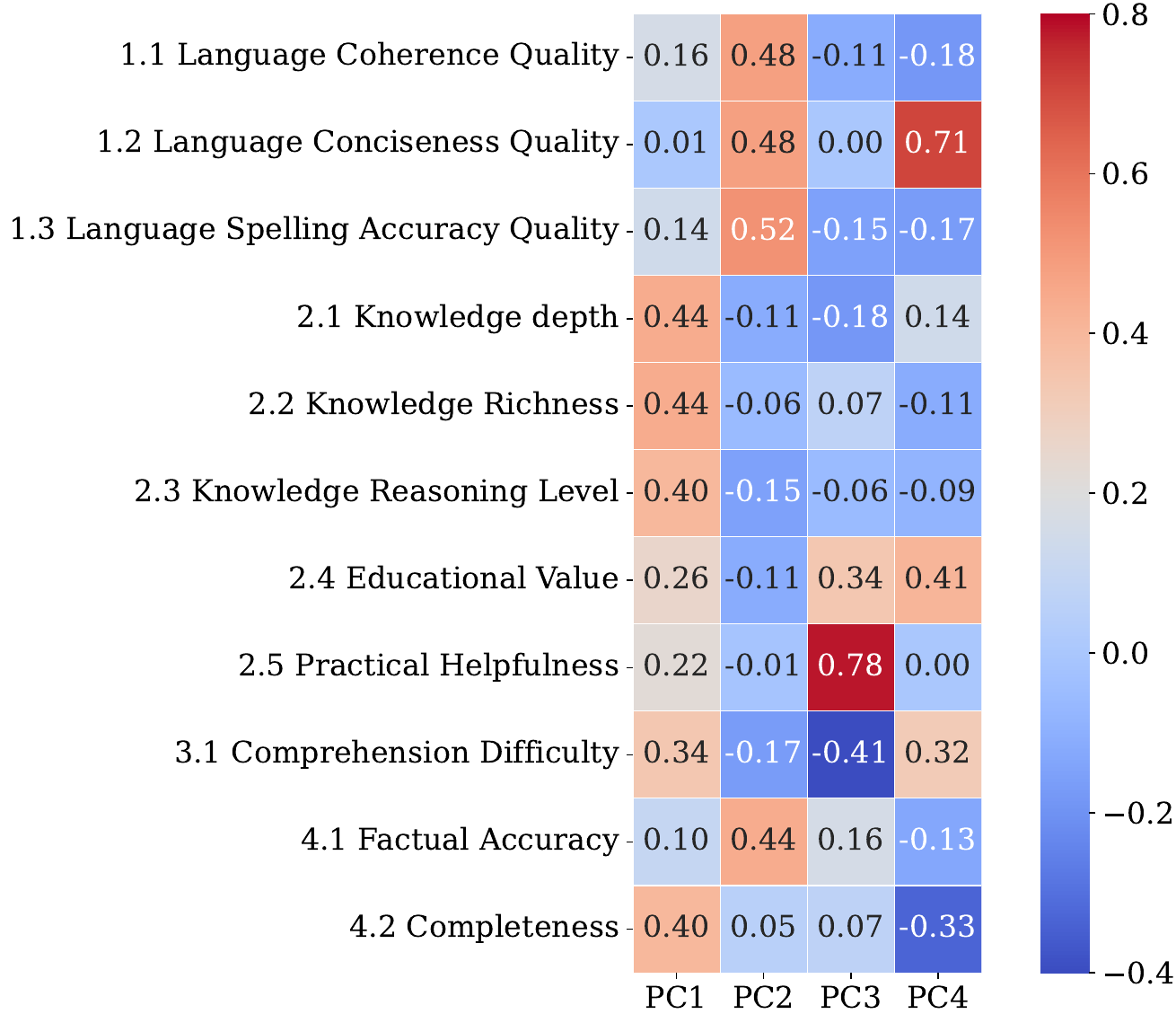}
        }
        \caption{Correlation analysis of evaluation dimensions. The correlations are calculated using 460k samples from the Fineweb-Edu dataset. Pairwise correlation coefficients are visualized as a heatmap.}
        \label{fig:correlation}
        \vspace{-0.5cm}
    \end{figure}

\textbf{Correlation of original dimensions.} Figure \ref{fig:correlation_matrix} describes the correlation coefficient across 11 dimensions with scores using 460k examples from Fineweb-Edu dataset. We observe that most dimensions exhibit weak correlation (correlation coefficient $<0.5$), suggesting that the metrics capture distinct aspects of the data. Nonetheless, we can still discover moderate correlation between the dimensions, such as knowledge depth vs. knowledge richness, and completeness vs. knowledge richness, indicating partial overlaps in their coverage. Moreover, nearly all the dimensions exhibit at least some degree of correlation with one another. These results suggest that although different metrics may appear conceptually orthogonal from a human perspective, they are not strictly independent in practice or from the model's viewpoint, validating the necessity of dimension decomposition.

\textbf{Correlation between original dimensions and principal components.} After applying PCA to transform dimensions into orthogonal principal components, Figure \ref{fig:correlation_matrix_pc} describes the correlations between PC dimensions and original dimensions. Each PC exhibits a strong correlation with a subset of the original dimensions, indicating that they can represent meaningful information or semantic features. For example, PC1 aligns strongly with knowledge quality and comprehension difficulty, highlighting its central role in characterizing overall data quality. By contrast, language-related dimensions exhibit weaker direct alignment with the leading PCs. This observation indicates that the linguistic factors may already be embedded within other correlated dimensions, e.g., knowledge quality, and thus they do not emerge as dominant signals in the first few components. More generally, the orthogonal principal components are linear combinations of the original dimensions. They should not be interpreted as single semantic dimensions, but rather as abstract features that combine multiple correlated attributes and capture salient aspects of the dataset from the model’s perspective.

\subsubsection{Orthogonality Between Dimensions}
\begin{figure}[h]
    \centering
    \vspace{-0.3cm}
        \subfloat[UMAP for embeddings. The embeddings are generated through 1000 sampled top-scored data from each PC dimension.]{
            \centering
            \label{fig:embedding}
            \includegraphics[width=0.30\linewidth]{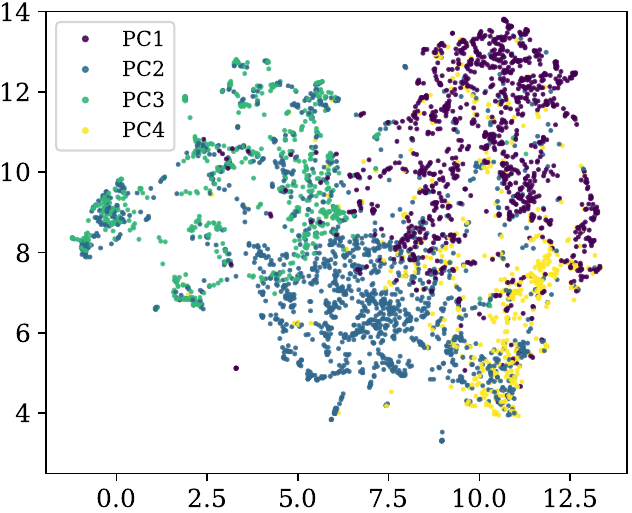}
        }
        \hspace{0.5cm}
        \subfloat[UpSet plot of top-scored subsets. Data selected with top scores of each dimension are shown, with intersections indicating data that simultaneously belong to multiple top-scored subsets.]{
            \centering
            \label{fig:upset-graph}
            \includegraphics[width=0.46\linewidth]{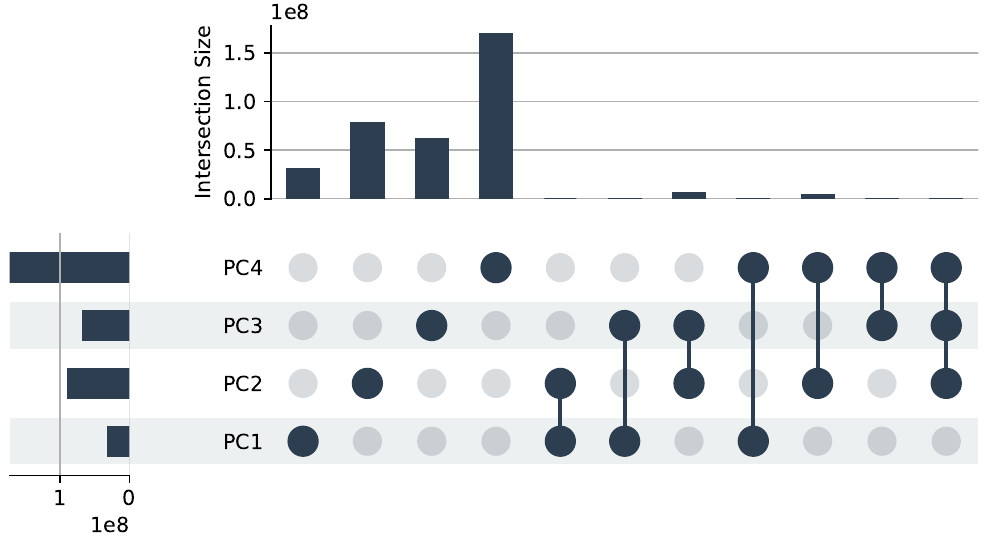}
        }
        \caption{Data orthogonality from different PC dimensions. The data is top-scored tokens from each PC dimension, which constitutes a 100B tokens dataset.}
        \label{fig:dimension-orthogonality}
        \vspace{-0.5cm}
    \end{figure}

To assess the distinctiveness of different PC dimensions, we visualize the embeddings of top-scored data from each dimension using UMAP projection and an Upset plot, as shown in Figure \ref{fig:dimension-orthogonality}. The UMAP visualization in Figure \ref{fig:umap-after} reveals that samples from different dimensions occupy separable regions, which validates the orthogonality between dimensions and suggests that combining data from different dimensions can yield complementary data subsets. Figure \ref{fig:upset-graph} illustrates the marginal intersection of the data from different PC dimensions, with an overlapping ratio of 2\% after tokenization, further confirming the effectiveness of the dimension decomposition.

\subsubsection{Scaling with More Dimensions} 
To examine whether increasing the number of PCs improves performance, we conduct an ablation study with varying PC dimensions, as summarized in Table~\ref{tab:ablation}. The results show that the performance improves steadily up to four dimensions, after which the performance gain becomes saturated. Beyond four PCs, additional dimensions contribute marginally to the model performance, likely because lower-variance PCs contain less information or noisy signals. Additionally, selecting more PCs will introduce increased computational cost, as labeling a large corpus is both computationally and time-consuming. These findings suggest that a small set of carefully selected orthogonal dimensions is sufficient for robust data selection, striking a balance between efficiency and effectiveness.
\begin{table}[h]
\centering
\begin{tabular}{l|C{1.4cm} C{1.4cm} C{1.4cm} C{1.4cm} C{1.4cm} C{1.4cm}}
\toprule
\textbf{Method}  &\textbf{Arc-C} &\textbf{Arc-E} &\textbf{Hellaswag} &\textbf{SIQA} &\textbf{PIQA} &\textbf{Average}
\\ \midrule
PC1 & 0.3635 & 0.6035 & 0.4309 & 0.811 & 0.6436 & 0.5705 \\ 
PC1-2 & 0.3575 & 0.6107 & 0.5372 & 0.830  & 0.7514 & 0.6174 \\
PC1-3 & 0.3635 & 0.6090 & 0.5936 & 0.791 & 0.7748 & 0.6264 \\
PC1-4 & 0.4053 & 0.6570 & 0.5741 & 0.872  & 0.7715 & 0.6559 \\
PC1-5 & 0.3938 & 0.6692 & 0.5689 & 0.880 & 0.7669  & 0.6557 \\
PC1-6 & 0.4249 & 0.7101 & 0.5606 & 0.872 & 0.7598  & 0.6654 \\
\bottomrule
\end{tabular}
\caption{Performance across varying PC numbers. The data budget is set at 100B tokens for all experiments, and we evenly distribute the budget across each dimension.}
\vspace{-0.3cm}
\label{tab:ablation}
\end{table}

\section{Related Works}
With the increasing scale of both model and corpora size, there is a growing demand for efficient methods to select high-quality pretraining data. The quality and diversity are two key considerations during data selection. Existing data selection methods have been proposed to enhance data quality and diversity through three main directions: non-classifier-based methods, single-classifier-based methods, and multi-classifier-based methods. 

\textbf{Non-classifier-based methods.} Various works have relied on rule-based filtering with explicit heuristics or deterministic criteria (\cite{laurenccon2022bigscience, weber2024redpajama, penedo2023refinedweb, raffel2020exploring, lee2021deduplicating}), including language identification, URL blocks, content de-duplication, and document length thresholds. Moreover, these approaches are combined into multi-stage pipelines that sequentially perform cleaning, deduplication, quality, and safety filtering (\cite{nguyen2023culturax}). 
While rule-based methods effectively improve data quality and reduce noisy data, they fail to inspect semantic-level information and will introduce distribution bias.

\textbf{Single-classifier-based methods.} In contrast to the rule-based method, it utilized a learned scoring function or discriminator to label and filter out high-quality data (\cite{wenzek2019ccnet, touvron2023llama, wettig2024qurating, penedo2024fineweb, su2024nemotron, wang2025ultra}). Among them, language modeling perplexity has been adopted to identify data with high quality (\cite{wenzek2019ccnet, touvron2023llama}). Methods like QuRating (\cite{wettig2024qurating}) and FineWeb-Edu (\cite{penedo2024fineweb}) utilize classifiers that focus on specific aspects of LLM capabilities, such as reading comprehension and knowledge acquisition. Moreover, works like Ultra-FineWeb (\cite{wang2025ultra}) and DSIR (\cite{xie2023data}) utilize a target dataset to guide the classifier in predicting the quality of the data. Although methods with a single classifier can effectively filter out data with certain desirable features, their reliance on a single evaluation dimension limits data diversity and often leads to imbalanced capabilities in the trained models.

\textbf{Multi-classifier-based methods.} More recent works have attempted to incorporate multi-dimensional evaluation during data selection. Compared with single-dimensional methods, the combination of multi-dimensional classifiers can provide a more comprehensive evaluation of data. However, it remains a challenge to balance the influence from different dimensions. One line of the research typically assigns weights to each dimension through performance tests on small proxy models (\cite{zhuang2025meta, bai2025efficient}). However, the dimensional correlations are not well-addressed, resulting in bias in the combined scores and reduced data diversity. Another line of the research design sampling ratios for different domains (\cite{liu2025quadmix}) to ensure data diversity. However, the inherent overlapping of the domains will still lead to bias in the selected data. As a result, the traditional top-$k$ method is ineffective in the multi-dimensional setting, leaving the question of integrating multi-dimensional evaluations open.   

\section{Conclusion}
In this work, we investigated the underlying cause of bias in score-based data selection and identified that neglecting diversity leads to non-monotonicity between the dataset scores and model performance. To address this issue, we proposed ODiS, a method that explicitly mitigates the correlation between different data features while ensuring data diversity and retaining high-quality data. The experiment results demonstrated that ODiS can effectively mitigate inter-dimensional correlation, enhance data diversity, and consistently improve model performance across various downstream tasks compared to several baselines. These findings highlight that effective data selection for pre-training models should consider quality and diversity jointly, and the correlation between different dimensions must be addressed appropriately. Looking forward, we encourage future data selection works to consider the neglected diversity as a cause of performance degradation and adopt appropriate measures to enhance data diversity. 

\newpage

\section*{Ethics Statement}
The authors confirm that this work adheres to the ICLR Code of Ethics. Our research was conducted in accordance with recognized ethical standards, and we have carefully examined the societal, environmental, and potential misuse implications of our contributions. 

\section*{Reproducibility Statement}
The authors have made extensive efforts to ensure the work's Reproducibility, including datasets, evaluation metrics, methodology, models. 

The details of the datasets used in this work are all open-sourced, and we have described them in the Section \ref{sec:experiment-setup}. The evaluation metrics and prompt for the dimensions can be found in Appendix \ref{appd:11-dimension} and Appendix \ref{appd:score-metric}. The details of the evaluation benchmarks are described in Section \ref{sec:experiment-setup}. The proposed method is described in detail, and the pseudocode is provided. All the implementation details are provided during the description. We utilized the LLaMA-3 model as the base model and adjusted the parameters to obtain a 1.5B-parameter model for training. Besides, the Roberta model can be obtained on the HuggingFace website. All the experimental results are reproducible, and we have averaged the results from multiple experiments to ensure an accurate result. 

We believe these detailed descriptions are sufficient to reproduce our results.

\bibliography{iclr2026_conference}
\bibliographystyle{iclr2026_conference}

\newpage
\appendix

\section{Use of Large Language Models}
During manuscript preparation, we occasionally utilized large language models (LLMs) to refine language expression, such as improving sentence fluency, and enhancing readability. The model was not involved in generating original research contributions, including research direction formulation, methodologies selection, experiment designs, results analysis. All the core intellectual work, such as idea development, experiment execution, and results interpretation, was carried out independently by the authors. Any linguistic suggestions offered by the LLM were carefully reviewed and selectively incorporated, ensuring that accuracy, originality, and scholarly integrity were fully maintained. The authors alone take responsibility for the research content and conclusions, and the LLM is not listed as a contribution or author.

\section{11 evaluation dimensions \label{appd:11-dimension}}
To comprehensively evaluate the quality and usefulness of training data, we design an 11-dimensional evaluation framework. These dimensions aim to capture complementary aspects of data that jointly determine its contribution to pretraining large language models (LLMs). Specifically, the dimensions are grouped into four general categories: language quality, which reflects the clarity and fluency of expression; knowledge quality, which measures the depth, diversity, and utility of information; comprehension difficulty, which reflects the complexity of content and its potential to improve generalization; and information quality, which ensures factual correctness and completeness. Together, these dimensions provide a multi-perspective evaluation of data, enabling a balanced assessment of both quality and diversity. Below, we provide the details of each category:
\begin{enumerate}
    \item \textbf{Language quality:} LLMs fundamentally rely on languages to understand the content and interact with humans. High-quality data contributes directly to improving models’ comprehension and generation abilities. A high-quality document should present ideas in a logical and rational organization, avoid redundant and irrelevant content, and use accurate spelling and grammar to convey meaning (\cite{penedo2023refinedweb}). To capture these properties, we evaluate language quality in three dimensions: (i) \emph{coherence}, which reflects whether the text follows a logical and consistent structure; (ii) \emph{conciseness}, which measures whether the information is conveyed efficiently without unnecessary repetition; and (iii) \emph{spelling/grammar accuracy}, which ensures correctness in word usage and sentence construction.  
    
    \item \textbf{Knowledge quality:} Beyond linguistic clarity, high-quality data must provide comprehensive knowledge to enrich an LLM’s understanding of the world. 
    This dimension mainly measures whether a document contains valuable, diverse, and practical information that can improve the model’s reasoning ability and enhance the factual knowledge base (\cite{gunasekar2023textbooks}). Recent work has also shown that the small models can approach the performance of larger ones if trained with reasoning data (\cite{guo2025deepseek}). To comprehensively capture knowledge quality, we define five dimensions: (i) \emph{knowledge depth}, assessing the extent to which a document explores concepts beyond superficial descriptions; (ii) \emph{knowledge richness}, measuring the breadth of covered topics and perspectives; (iii) \emph{reasoning}, capturing the presence of explicit logical inference, argumentation, or step-by-step derivations; (iv) \emph{educational value}, evaluating whether the content provides clear, structured explanations suitable for learning or instruction; and (v) \emph{practical helpfulness}, which assesses the applicability of the knowledge to real-world problems or everyday use.

    \item \textbf{Comprehension difficulty:} Challenging and complex data can enhance better generalization and adaptability in LLMs. Texts with higher conceptual complexity, specialized domain knowledge, or multi-step expository structures push models to develop deeper comprehension abilities and to generate more professional and domain-specific responses \cite{agrawal2023corpus}. This dimension therefore evaluates the difficulty of the data, considering factors such as the abstractness of concepts, the level of technicality or professionalism, and the requirement for multi-stage reasoning.

    \item \textbf{Information Quality: } For models to learn accurate and reliable representations, it is crucial that the training data provides factual, complete, and unambiguous information. Documents with inaccurate or incomplete facts risk propagating errors and degrading downstream performance (\cite{chang2024large}). To evaluate this aspect, we define two evaluation dimensions: (i) \emph{factual accuracy}, which measures the degree to which the document presents correct and verifiable information; and (ii) \emph{completeness}, which reflects whether the information is sufficiently detailed and covers all key aspects of the described topic, thereby reducing the risk of the model learning partial or misleading facts.
\end{enumerate}

\section{Metrics and Prompt for each dimension \label{appd:score-metric}}
To evaluate data quality and diversity across multiple dimensions, we design a set of prompts that guide large language models to score documents on 11 distinct dimensions, as described in Appendix \ref{appd:11-dimension}. Each dimension is assessed on a tailored Likert scale ranging from 0 to 3/4/5 points depending on the property being measured, with detailed criteria provided for each score level. The prompts are designed to generate both a quantitative score and a brief qualitative justification, ensuring transparency in the evaluation process. These scores are then aggregated into the score matrix used in our PCA-based data selection framework (see Section~\ref{sec:multi-dimensions}).

\begin{lstlisting}[language={python},caption={Prompt for Coherence}]
Below is an extract from a web page. Evaluate whether the text demonstrates high coherence in terms of language quality. Please follow the following guideline to assess the language quality of the given extract on a 4 likert scale:

0 Point: Incomprehensible
- The text is grammatically chaotic and difficult to understand.
- Severe errors in structure, agreement, and tense prevent understanding.

1 Point: Partially Readable
- Some sentences are clear, but overall clarity is lacking.
- Noticeable grammatical errors and inconsistency disrupt smooth reading.

2 Points: Moderately Coherent
- Occasional language issues but overall understandable.
- Logical flow is maintained with some awkward phrasing.

3 Points: Generally Coherent with Minor Errors
- Paragraphs progress logically with minor, infrequent language errors.
- Sentences are generally well-formed with consistent tense and clear subject-verb agreement.

4 Points: Exceptionally Coherent
- The text is grammatically flawless, with precise subject-verb agreement and tense usage.
- Sentence and paragraph structure is logically ordered and fluid.
- Punctuation and syntax enhance the clarity and flow of ideas.

The extract:  {text}  
After examining the extract:
 - Briefly justify your total score, up to 50 words.  
 - Conclude begin with the score using the format: "Language Coherence Score: <total points>"
\end{lstlisting}

\begin{lstlisting}[language={python},caption={Prompt for Conciseness}]
Below is an extract from a web page. Evaluate whether the text demonstrates a high level of conciseness. Please follow the following guideline to assess the conciseness of the given extract on a 4 likert scale:

0 Point: Excessively Wordy
- The extract is filled with redundant, unrelated, or repetitive language.
- Nearly every sentence could be significantly shortened or removed without loss of meaning.
- Core ideas are obscured or lost in verbosity.

1 Point: Somewhat Wordy
- The text is clear but contains noticeable repetition or unnecessary words.
- Some sentences are overly elaborate.

2 Points: Moderately Concise
- The extract avoids major redundancy but may include some unnecessary elaboration.
- Most sentences convey meaning efficiently, though small improvements in brevity are possible.
- The main points are clear and not lost in superfluous language.

3 Points: Concise and Effective
- Ideas are expressed clearly and directly, with minor redundancy or unnecessary details.
- Minimal to no repetition or fluff.

4 Points: Exceptionally Concise
- Every word is essential and contributes directly to the meaning.
- No repetition, filler, or unnecessary elaboration.
- The writing is focused, impactful, and efficient.


- The extract:  {text}  

After examining the extract:
 - Briefly justify your total score, up to 50 words.  
 - Conclude begin with the score using the format: "Language Conciseness Score: <total points>"
\end{lstlisting}

\begin{lstlisting}[language={python},caption={Prompt for Spelling Accuracy}]
Below is an extract from a web page.  Evaluate whether the text demonstrates high accuracy of word usage, which contributes to the as overall language quality. Please follow the following guideline to assess the accuracy of word usage in the given extract on a 4 likert scale:

0 Points: Severe Inaccuracy
- The extract contains frequent incorrect word usages.
- Frequent typos, incorrect word forms, or misuse of words make the text almost unreadable.
- Errors severely hinder understanding.

1 Points: Limited Accuracy
- Spelling mistakes appear regularly but are not overwhelming.
- Occasional misuse of words or minor typos affect clarity.
- The overall message is still understandable but occasionally unclear.

2 Points: Moderate Accuracy
- Most of the text is correctly spelled, with some minor errors or infrequent typos.
- Occasional confusion between similar-sounding words may appear but does not significantly affect meaning.
- The extract remains mostly readable and understandable.

3 Points: Strong Accuracy
- Spelling is generally correct throughout.
- Only rare, minor typos or homophone errors are present, and they do not interfere with comprehension.
- The extract demonstrates clear attention to written accuracy.

4 Points: Perfect Accuracy
- The extract is free from any spelling errors, typos, or homophone confusion.
- All words are used appropriately and are correctly spelled.
- The writing is polished and precise, reflecting excellent language control.

The extract:  {text}  

After examining the extract:
 - Briefly justify your total score, up to 50 words.  
 - Conclude begin with the score using the format: "Language Spelling Accuracy Score: <total points>"
\end{lstlisting}

\begin{lstlisting}[language={python},caption={Prompt for Knowledge Depth}]
Below is an extract from a web page. Evaluate whether the text demonstrates an appropriate depth of knowledge, particularly with regard to the grade level it targets. The following gudeline is used to assess whether a text has a high knowledge depth on a 5 likert scale:

0 Points: No Knowledge Depth
- The extract contains no meaningful or accurate knowledge.
- It lacks substance entirely and offers no educational value at any grade level.

1 Point: Shallow and Common Knowledge for Pre-K to Grade 1
- The content is understandable even to early primary grades (Pre-K to Grade 1).
- Contain simple, basic facts or common knowledge (e.g., basic facts like "grass is green" or "2 + 2 = 4").

2 Points: Basic Knowledge for Lower Grades (Grades 2-4)
- The content is at lower elementary levels.
- Introduces simple concepts and provides very short, basic explanations.
- Requires understanding of simple definitions and explicit information.

3 Points: Introductory Knowledge for Middle Grades (Grades 5-7)
- Understandable for upper elementary to early middle school.
- Explains foundational concepts with some detail and structure.
- Some depth is present. It may require understanding of cause-and-effect relationships and ability to follow multi-step explanations.

4 Points: Substantive Knowledge for Secondary Levels (Grades 8-12)
- Content is well-developed and appropriate for high school.
- Explores concepts in depth, including underlying principles, reasoning, and potential implications.
- Characterized by complex sentence structures, theoretical concepts, evidence or examples to support points; resembles textbook content.

5 Points: Advanced Knowledge Depth (college-level or graduate-level)
- The extract reflects college-level or graduate-level understanding.
- The knowledge is usually only known to the professional people in a certain field. 
- May presents complex information, including detailed analysis, theoretical frameworks, multiple perspectives, and nuanced arguments.

The extract:  {text}  

After examining the extract:
 - Briefly justify your total score, up to 50 words.  
 - Conclude begin with the score using the format: "Knowledge Depth Score: <total points>"                       
\end{lstlisting}

\begin{lstlisting}[language={python},caption={Prompt for Knowledge Richness}]
Below is an extract from a web page. Evaluate whether the text demonstrates a high degree of knowledge density in its content. The following curriculum is used to assess whether a text has dense knowledge on a 4 likert scale:

0 Point: No Meaningful Knowledge
- The extract lacks any meaningful or specific content.
- No concrete facts, data, or identifiable concepts

1 Point: Minimal Knowledge Content
- Contains only 1-2 disjointed factual statements
- No context, sourcing, or explanation

2 Points: Moderately Knowledge Density
- The extract includes several points of useful knowledge.
- Support with some details, examples, or explanations.

3 Points: Substantially Rich in Knowledge
- The content provides a well-rounded and informative discussion.
- Ideas are explained with clarity and supported by relevant details or examples.

4 Points: Exceptionally Knowledge-Rich
- The extract offers a dense, nuanced, and well-connected presentation of knowledge.
- The content shows breadth and depth, encouraging comprehensive understanding.

The extract:  {text}  

After examining the extract:
 - Briefly justify your total score, up to 50 words.  
 - Conclude begin with the score using the format: "Knowledge Richness Score: <total points>"                    
\end{lstlisting}

\begin{lstlisting}[language={python},caption={Prompt for Reasoning Level}]
Below is an extract from a web page. Evaluate whether the text demonstrates a high level of reasoning level. The following curriculum is used to assess whether a text has a high reasoning level:

0 Points: No Reasoning Present
- The text lacks any evidence of thinking or reasoning from the writer.

1 Point: Minimal Reasoning
- Some claims are made, but reasoning is largely absent or extremely shallow.
- No causal relationships or inferential steps are evident.
- Readers are not encouraged to reflect or engage intellectually.

2 Points: Limited Reasoning
The text demonstrates some basic thinking and reasoning, such as: 
- a straightforward application of a known technique
- simple analysis of a problem. 

3 Points: Moderate Reasoning
The text demonstrates adequate level thinking and reasoning, such as 
- a consideration of multiple approaches to a problem. 
- A discussion of the trade-offs between different solutions. 

4 Points: Strong Reasoning
The text demonstrates significant thinking and reasoning, such as: 
- Multi-step reasoning chains to solve a complex problem.
- Advanced reasoning patterns often used in specialized science domains

5 Points: Exceptional Reasoning Quality
The text exemplifies exceptional thinking and reasoning, such as: 
- A highly innovative and creative approach to solving a complex problem in specialized domains. 
- Combining multiple reasoning and thinking techniques, with novel abstraction of the problem.

The extract:  {text}  

After examining the extract:
 - Briefly justify your total score, up to 50 words.  
 - Conclude begin with the score using the format: "Knowledge Reasoning Score: <total points>"
\end{lstlisting}

\begin{lstlisting}[language={python},caption={Prompt for Educational Value}]
Below is an extract from a web page. Evaluate whether the page has a high educational value for teaching from kindergarten to graduate education. The following curriculum is used to assess whether a text has a high educational value on a 3 point scale: 

**0 Point: No Educational Value**  
- Not even a single bit of information is worth learning. 
- Note that if there is even a single bit of information that is worth learning, the score should be at least 1 point.

**1 Point: Minimal Educational Relevance**  
- The extract provides some useful information pertinent that is worth learning or teaching, but does not align closely with educational standards. 
- It may include a large amount of non-educational content (e.g., advertisements, promotional material) that detracts from its usefulness.  

**2 Points: Suitable for Educational Use**  
- The extract provides a lot of useful information that is worth learning or teaching. The content is fluent and coherent.
- It may include a small amount of non-educational content. It may have limitations, such as incomplete coverage or extraneous information.

**3 Points: Highly Relevant and Beneficial**  
- The extract has very high educational value. It contains high density of information that is worth learning or teaching, either for any level of education.
- Content is clear, consistent, and focused, with minimal irrelevant information.
- May resemble a snippet from a textbook, tutorial, exercises, solutions, or any structured learning materials.

The extract: 
{text}

After examining the extract: 
- Briefly justify your total score, up to 50 words. 
- Conclude begin with the score using the format: "Educational score: <the assigned score>"
\end{lstlisting}

\begin{lstlisting}[language={python},caption={Prompt for Practical Helpfulness}]
Below is an extract from a web page.  Evaluate whether the content demonstrates a high degree of practical helpfulness, particularly in terms of offering applicable knowledge for real-world utility. The following curriculum is used to assess whether a text has a high practical helpfulness on a 4 likert scale:

0 Points: No Practical Helpfulness
- The extract contains no useful or applicable knowledge.
- May be purely entertainment or advertisement with zero actionable takeaways
- May contain misinformation or harmful suggestions

1 Point: Minor Utility
- The text may hint at applicable ideas but lacks clarity, specificity, or guidance.
- It is too general or abstract to be put into use.

2 Points: Moderately Helpfulness
- The knowledge can be applicable in some uncommon scenarios (targets <1% audience) that only relate to a small portion of people.

3 Points: Broadly Helpful
- The extract includes practical information that could be applied in common contexts.
- Offers validated strategies for common needs

4 Points: Substantially Helpful
- The extract offers clear, applicable knowledge or skills that are useful in real-world scenarios that frequently occur. 
- Addresses frequent pain points (>10% audience)


The extract:  {text}  

After examining the extract:
 - Briefly justify your total score, up to 50 words.  
 - Conclude begin with the score using the format: "Knowledge Practical Helpfulness Score: <total points>"       
\end{lstlisting}

\begin{lstlisting}[language={python},caption={Prompt for Comprehension Difficulty}]
Here is an extract from a webpage. Please evaluate the percentage of the global population that is likely to be able to comprehend the knowledge text. The following scale is used to assess the comprehension difficulty, with a 5-point Likert scale:

0 Points: No value to understand
- The content is incomprehensible due to its low language quality. 
- Contains gibberish, severe grammar errors, or formatting problems.
- Examples: Advertisement, machine-translated nonsense, corrupted text

1 Point: Universal Comprehension
- The content is very simple and direct, easily understood by the vast majority of people.
- Requies basic vocabulary (<4th grade level), commonsense knowledge, with no jargon.
- Examples: Weather reports, simple recipes, basic safety instructions

2 Points: Majority Effortless 
The content is clear and easily understandable for almost everyone, with only a very small percentage finding it difficult.
- Requires conversational language level and general world knowledge
- Examples: social media posts, most new articles

3 Points: Educated Majority
- The content is accessible to the majority of people, with some difficulty, but most people should be able to understand and comprehend it after some effort.
- Requires high school reading level and secondary education concepts
- Examples: Government pamphlets, workplace training manuals, simple financial advice. 

4 Points: Specialized Audience 
- The content is understood by a small portion of people, but it remains challenging for the majority.
- The content may require some expertise.
-  Requires undergraduate-level training in field
-  Examples: College textbooks, legal contracts, financial advice

5 Points: Expertise
- The content may be very professional or academic. 
- Requires graduate-level expertise.
- Examples: Quantum physics proofs, AI architecture patents, genomic research

Extract:
 {text}

After reviewing the text:
Briefly justify your total score in up to 50 words.
Conclude begin with the score using the format: "Comprehension Difficulty Score: "
\end{lstlisting}

\begin{lstlisting}[language={python},caption={Prompt for Factual Accuracy}]
Here is an extract from a webpage. Evaluate whether the content demonstrates a high level of factual accuracy as part of its overall information quality. 
Note that:
- the text may include some facts that are unknown to you. In these cases, you can ignore these unknown or uncertain facts and only focus on identify those obvious factual errors that are known to you. 
- In some special contexts, such as fictions, it is allowed to contain some imaginary facts. 


The following guideline is used to assess the factual accuracy, with a 3-point Likert scale:

0 Point: Evidently Inaccurate
- The extract is filled with incorrect information.
- Key claims are demonstrably wrong or contradict well-established facts.

1 Point: Highly Unreliable
- The extract contains multiple factual inaccuracies or distortions.
- Misleading phrasing or vague statements obscure the truth.
- While not entirely false, it cannot be trusted as a reliable source of information.

2 Points: Generally Accurate with Minor Issues
- <2 minor errors in peripheral details
- Occasional imprecise language without distorting meaning
- Preserves core truth despite technical imperfections

3 Points: Accurate and Trustworthy
- No detectable errors in verifiable claims.


Extract:
 {text}

After reviewing the text:
- Briefly justify your total score in up to 50 words.
- Conclude begin with the score using the format: "Information Factual Accuracy Score:"
\end{lstlisting}

\begin{lstlisting}[language={python},caption={Prompt for Completeness}]
Here is an extract from a webpage.  Evaluate whether the content demonstrates a high degree of completeness, specifically in terms of how fully the topic is covered and whether the information is presented with sufficient context. The following scale is used to assess the information completeness, with a 4-point Likert scale:


0 Point: Severely Incomplete
The extract offers only fragments of information or vague references to the topic.
Key background, definitions, or context are missing.
The presentation leaves readers with more questions than answers.

1 Point: Limited Completeness
The extract touches on parts of the topic but leaves significant gaps.
It may assume prior knowledge or skip necessary context.
Information is partial or unevenly distributed.

2 Points: Moderately Complete
The extract introduces the main topic and provides sufficient context to follow the discussion.
Some areas may be underdeveloped or missing, but overall understanding is possible.
It resembles a summary or introductory passage.

3 Points: Substantially Complete
The extract covers the topic in a well-rounded and balanced manner.
Most relevant aspects are addressed, with clear and sufficient context.
There may be minor omissions, but they do not disrupt comprehension.

4 Points: Exceptionally Complete
The extract thoroughly explores the topic with comprehensive coverage.
All necessary context is included, with no critical gaps.
It reflects a deep and well-structured presentation that anticipates and answers potential reader questions.

Extract:
 {text}
 
After reviewing the text:
Briefly justify your total score in up to 50 words.
Conclude begin with the score using the format: "Information Completeness Score: "
\end{lstlisting}

\newpage
\section{Benchmarks Selection \label{sec:benchmarks}}
\begin{figure}[h]
        \centering
        \includegraphics[width=0.9\linewidth]{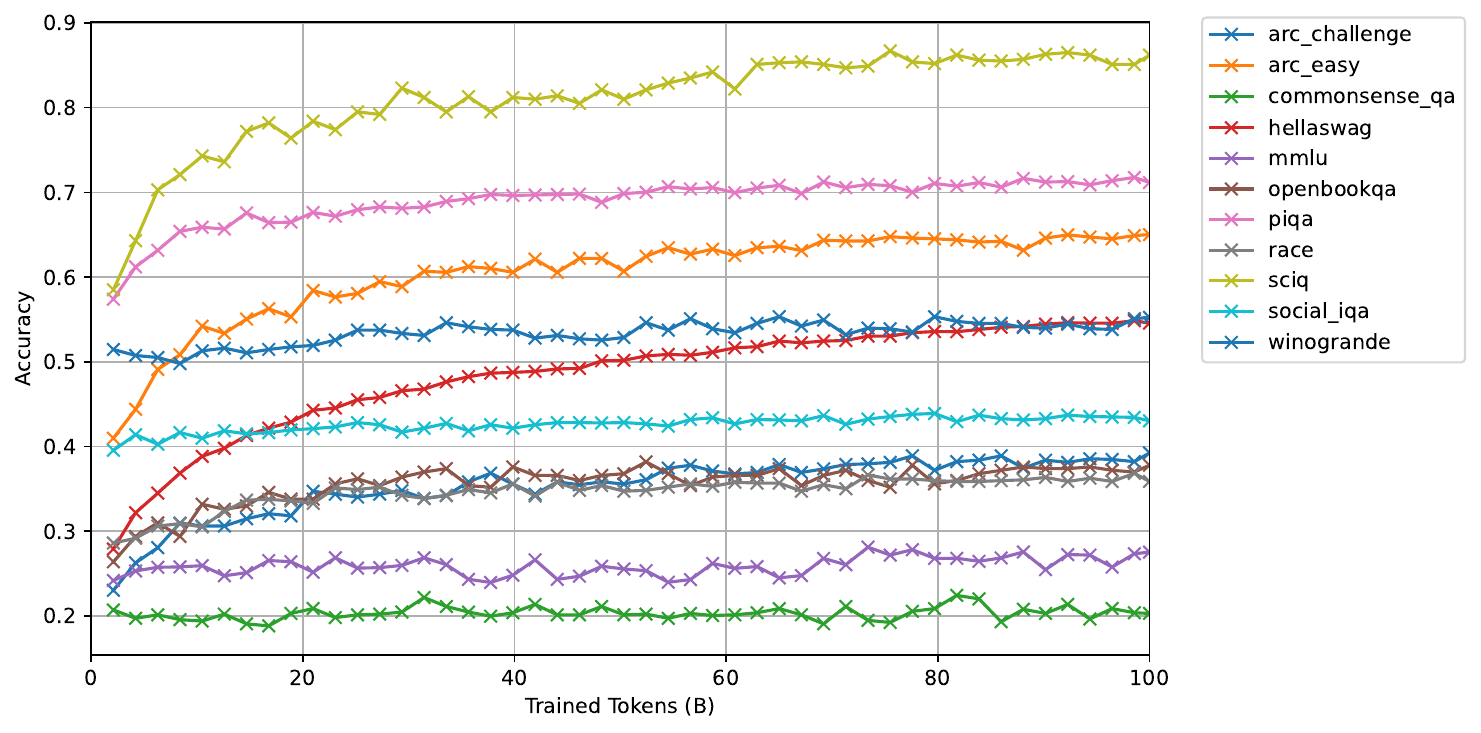}
        \caption{Performance across downstream tasks}
        \label{fig:performance-across-tasks}
    \end{figure}

To select appropriate downstream benchmarks that can effectively reflect the model performance, we observe the accuracy fluctuation as the trained data increases, with results reported in Figure \ref{fig:performance-across-tasks}. Arc-C, Arc-E, hellaswag, piqa, and sciq have obvious variation as the training progresses, while the rest of the benchmarks have a smaller performance improvement. Since our model and the trained data budget are relatively small, some benchmarks can not obviously reflect the training outcomes of the model. Therefore, we select Arc-C, Arc-E, hellaswag, piqa, and sciq as our benchmarks. 

\section{Score distribution across different PC}
\begin{figure}[h]
\centering
        \includegraphics[width=0.98\linewidth]{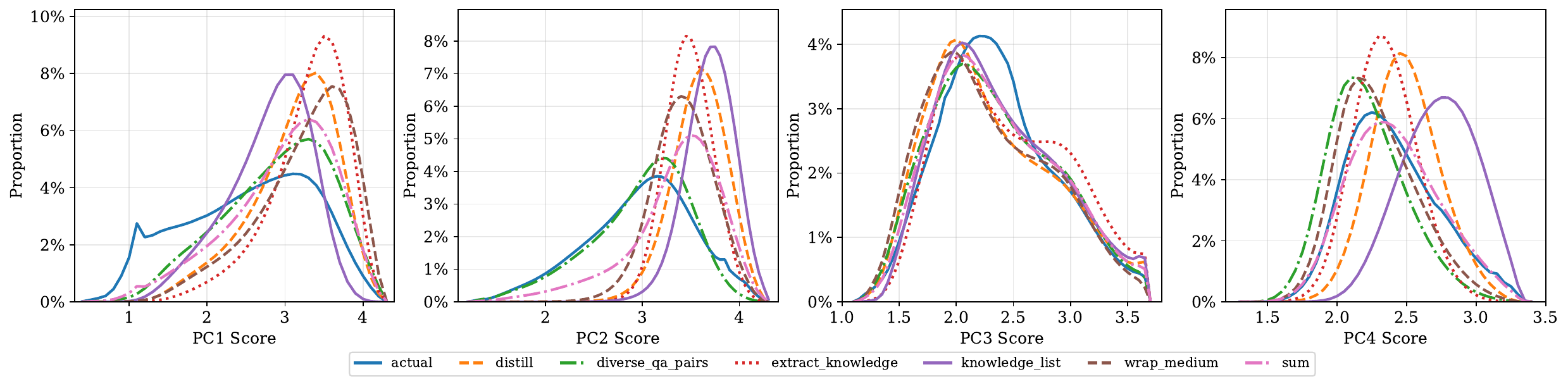}
    \caption{Score distribution over different domains.}
    \label{fig:distribution-pc}
\end{figure}

Figure \ref{fig:distribution-pc} demonstrates the score distribution over different PC dimensions on different domains. The domains are pre-devided by the Nemotron-CC dataset. We can observe that different PC dimensions emphasize distinct aspects, and joint selection across dimensions enhances data diversity. 

\newpage
\section{Results with single PC}
Table \ref{tab:single-pc} demonstrates the results of data selected with the single PC scorer. We can observe that each PC exhibits strength in certain area: PC1 and PC4 perform better on Arc-C and Arc-E, indicating a better ability at general knowledge, while PC2 and PC3 perform better on Hellaswag and PIQA, indicating a better ability for commonsense and physical reasoning. Moreover, models trained with top-scored data from each PC dimension consistently underperform, while sampling from a larger score range enhances the performance. These results highlight that different PC scorers focus on distinct data features and using one of them alone can not achieve the best performance.

\begin{table}[ht]
\centering
\begin{tabular}{l|C{1.4cm} C{1.4cm} C{1.4cm} C{1.4cm} C{1.4cm} C{1.4cm}}
\toprule
\textbf{Method}  &\textbf{Arc-C} &\textbf{Arc-E} &\textbf{Hellaswag} &\textbf{SIQA} &\textbf{PIQA} &\textbf{Average}
\\ \midrule
PC1-top & 0.3635 & 0.6035 & 0.4309 & 0.811 & 0.4289 & 0.5705 \\ 
PC2-top & 0.3072 & 0.5097 & 0.5567 & 0.727  & 0.4483 & 0.5707 \\
PC3-top & 0.3311 & 0.5551 & 0.6178 & 0.741 & 0.4386 & 0.6059 \\
PC4-top & 0.4053 & 0.7041 & 0.4484 & 0.879 & 0.4350  & 0.6245 \\
PC1-Sample & 0.3951	& 0.6519 & 0.5464 & 0.863 & 0.7116 & 0.6336\\
PC2-Sample & 0.3686	& 0.6103 & 0.5759 & 0.765 & 0.7448 & 0.6129\\
PC3-Sample & 0.3686	& 0.6557 & 0.6112 & 0.846 & 0.7579 & 0.6479\\
PC4-Sample & 0.4087	& 0.6860 & 0.5356 & 0.861 & 0.7318 & 0.6446\\
\bottomrule
\end{tabular}
\caption{Performance across PC dimensions.}
\label{tab:single-pc}
\end{table}

% \begin{table}[t]
% \centering
% \caption{Sample table title}
% \label{tab:main_results}
% \begin{tabular}{>{\centering\arraybackslash}p{2.7cm}|>{\centering\arraybackslash}p{1.4cm}>{\centering\arraybackslash} p{1.4cm}>{\centering\arraybackslash} p{1.4cm}>{\centering\arraybackslash} p{1.4cm}>{\centering\arraybackslash} p{1.4cm}>{\centering\arraybackslash} p{1.4cm}}
% \toprule
% \textbf{Data selection Method}  &\textbf{Arc-Challenge} &\textbf{Arc-Easy} &\textbf{Hellaswag} &\textbf{SIQA} &\textbf{PIQA} &\textbf{Average}
% \\ \midrule
% Random Selection & 0.3503 & 0.6273 & 0.5825 & 0.4537 & 0.7448 & 0.6320 \\ 
% 100B & 0.3532 & 0.5791 & 0.5332 & 0.4222 & 
% 0.7448 & 0.5859  \\ 
% 200B & ~ & ~ & ~ & ~ & ~ & ~  \\ 
% 300B & 0.378 & 0.6288 & 0.5754 & 0.4381 & 0.7552 & 0.6237 \\ 
% 400B & 0.3942 & 0.6153 & 0.5832 & 0.4263 & 0.7579 & 0.6321  \\ 
% 500B & 0.3823 & 0.6427 & 0.5832 & 0.4371 & 0.7476 & 0.6338  \\ 
% 600B & ~ & ~ & ~ & ~ & ~ & ~  \\ 
% 700B & 0.3814 & 0.6402 & 0.5818 & 0.4555 & 0.7465 & 0.6362  \\ 
% 800B & 0.3848 & 0.6507 & 0.5858 & 0.436 & 0.7503 & 0.6391  \\ 
% 900B & 0.4104 & 0.6692 & 0.5896 & 0.4442 & 0.7622 & 0.6545  \\ 
% Nemotron-CC HQ & 0.3725 & 0.6463 & 0.5774 & 0.4519 & 0.7356 & 0.6341  \\ 
% \bottomrule
% \end{tabular}
% \vspace{0.2cm}
% \parbox{\linewidth}{
% \footnotesize
% \textit{Sample table title}
% }
% \end{table}

\end{document}